\documentclass[conference]{IEEEtran}
\IEEEoverridecommandlockouts                              
\usepackage{graphicx}
\usepackage{tabularx}
\usepackage[utf8]{inputenc}
\usepackage[T1]{fontenc}
\usepackage{subcaption}

\newsavebox{\twosubbox}
\usepackage{xcolor, hyperref}
\usepackage{color,soul}
\def\BibTeX{{\rm B\kern-.05em{\sc i\kern-.025em b}\kern-.08em
    T\kern-.1667em\lower.7ex\hbox{E}\kern-.125emX}}

\begin{document}

\title{Co-speech Gestures for Human-Robot Collaboration
\thanks{Project funding was received from EU's Horizon 2020 research and innovation programme, grant no. 871449 (OpenDR).}
}

\author{\IEEEauthorblockN{Akif Ekrekli}
\IEEEauthorblockA{\textit{Engineering and}\\\textit{Natural Sciences} \\
\textit{Tampere University}\\
Tampere, Finland \\
akif.ekrekli@tuni.fi}
\and
\IEEEauthorblockN{Alexandre Angleraud}
\IEEEauthorblockA{\textit{Engineering and}\\\textit{Natural Sciences}  \\
\textit{Tampere University}\\
Tampere, Finland \\
alexandre.angleraud@tuni.fi}
\and
\IEEEauthorblockN{Gaurang Sharma}
\IEEEauthorblockA{\textit{Engineering and}\\\textit{Natural Sciences} \\
\textit{Tampere University}\\
Tampere, Finland \\
gaurang.sharma@tuni.fi}

\and
\IEEEauthorblockN{Roel Pieters}
\IEEEauthorblockA{\textit{Engineering and}\\\textit{Natural Sciences} \\
\textit{Tampere University}\\
Tampere, Finland \\
roel.pieters@tuni.fi}
}

\maketitle

\begin{abstract} 
Collaboration between human and robot requires effective modes of communication to assign robot tasks and coordinate activities. As communication can utilize different modalities, a multi-modal approach can be more expressive than single modal models alone. In this work we propose a co-speech gesture model that can assign robot tasks for human-robot collaboration. Human gestures and speech, detected by computer vision and speech recognition, can thus refer to objects in the scene and apply robot actions to them. We present an experimental evaluation of the multi-modal co-speech model with a real-world industrial use case. Results demonstrate that multi-modal communication is easy to achieve and can provide benefits for collaboration with respect to single modal tools.
\end{abstract}
\section{Introduction}
\begin{figure}[t]
    \centering
    \includegraphics[width=\linewidth]{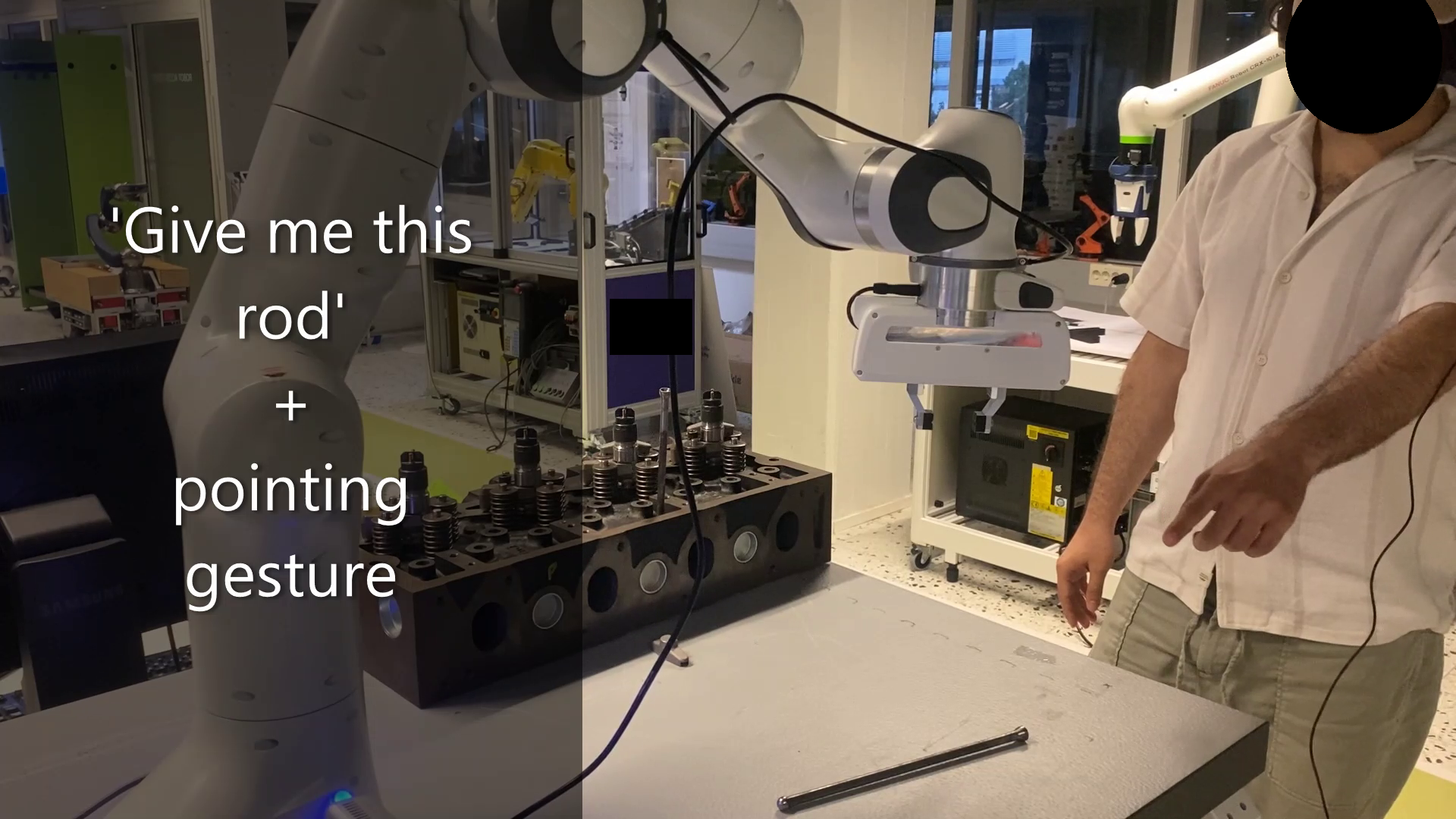}
    \caption{Co-speech gesture model that combines a speech phrase, human gesture detection and object perception to command robot actions.
    }\label{fig:motivation}
\end{figure}

Fluent interaction between human and robot requires reliable perception to capture the commands of a person. While recent approaches in deep learning \cite{Robinson2022} have established impressive tools to detect e.g., human pose, gestures and speech, single tools alone can not always convey easily the commands intended \cite{gross2023communicative}. Reasons for this are the limited expressions available for different modes of communication and the limitations in perception performance. Human hand gestures, for example, contain much less information content than speech. On the other hand, gesture detection can be done much quicker than speech recognition, leading to a faster response time. These conflicting properties motivate to combine multiple perception tools into a single multi-modal detection model that utilizes communication from human to robot for assigning tasks and coordinating the collaboration. In this work we compare different perception tools and analyse them with respect to their suitability for human-robot collaboration. A co-speech gesture model is then developed that combines speech, human hand gestures and object detection to achieve effective communication of desired robot tasks, such as picking human-specified objects and robot to human hand-overs (see Fig. \ref{fig:motivation}). The developments are intended for industrial human-robot collaboration where a collaborative robot shares its tasks, and works in close collaboration with, a human operator.
Our contributions are:

\begin{itemize}
    \item Human speech and hand gesture perception methods to command robot actions
    \item Co-speech gesture model that combines human natural speech and hand gestures to command robot actions
    \item Experimental evaluation of the co-speech gesture model in an industrial human-robot collaborative use case
\end{itemize}


\section{Related work}

\subsection{Human-Robot Collaboration}

Collaboration between human and robot is often targeted for industrial manufacturing \cite{Villani2018}, as both robot and human have unique skills that complement each other. Different interfaces that enable the collaboration have been analyzed, providing clear directions on how the collaboration benefits the tasks \cite{Wang2019}. Approaches include voice processing, gesture recognition, haptic interaction, and even brainwave perception. Often machine \cite{Semeraro2023} and deep \cite{Sunderhauf2018} learning are used as enabling perception tool \cite{Robinson2022} to classify and recognize the person and objects in the environment \cite{Fan2022}. 

\begin{figure*}[t]
    \centering
    \includegraphics[width=0.8\linewidth]{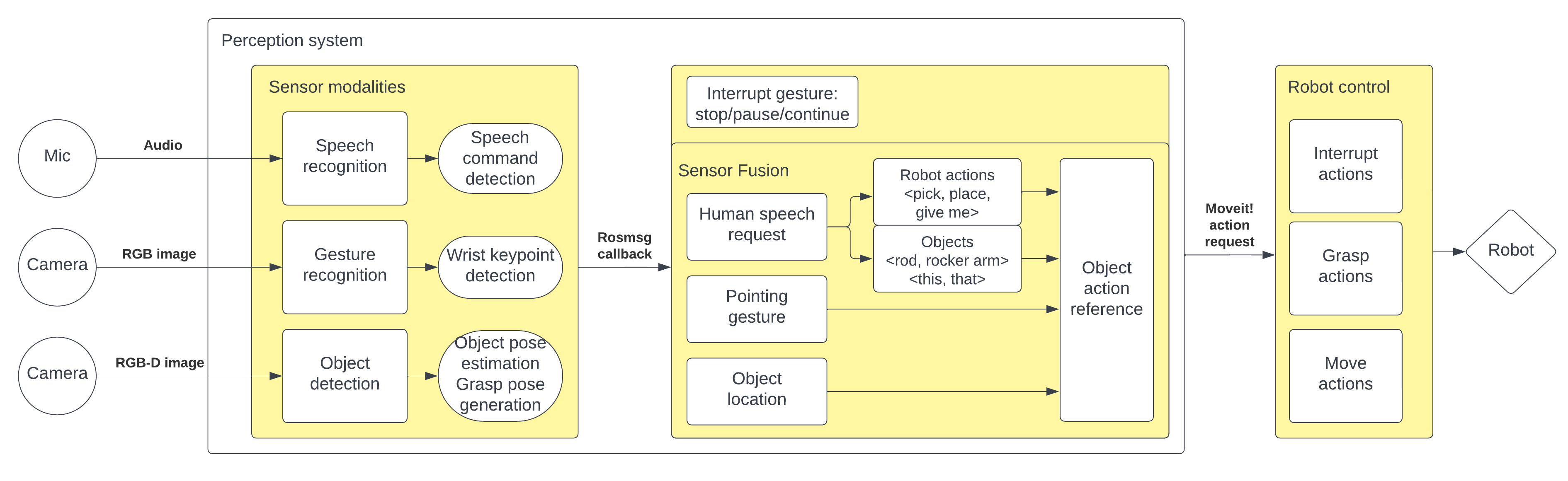}
    \caption{Co-speech gesture model that takes input from speech commands, gesture recognition and object detection to generate robot actions for human-robot collaboration. Sensor fusion enables the human to refer to specific objects (\texttt{<rod, rocker, arm, this, that>}) and apply actions to them (\texttt{<pick, place, give me>}).
    }\label{fig:co-speech gesture model}
\end{figure*}

\subsection{Human Perception}

Visual detection of a person in the scene has been an active area of research \cite{Zacharaki2020}.
Different visual modalities have been utilized \cite{Linder2021}, such as RGB and depth information \cite{Magrini2020}. Multi-modal approaches that utilize RGB-D data are popular as well \cite{Qi_2018_CVPR}. Human pose estimation goes a step further than human detection by estimating the 3D pose of a human and their individual skeleton joints \cite{Osokin2018}, which can be used as input for gesture detection.
Utilizing speech for commanding robots has been demonstrated with short verbal commands for task coordination \cite{Angleraud2021} and task programming \cite{ionescu2021programming}. As extension to short speech commands, natural language as instructions to robots has been used for planning \cite{boteanu2016model} and allocation \cite{behrens2019specifying} of tasks to be performed by the robot. 

Multi-modal human-robot collaboration using gestures and speech simultaneously has been demonstrated for a human interacting with the humanoid robot NAO in \cite{bremner2015efficiency}, where short phrases and gestures are utilized to indicate human actions.
Collaboration between a robot arm and a human worker is also demonstrated in \cite{chen2022}, where a set of gestures and speech commands are perceived individually to produce the same input for robot actions.
As comparison, our work considers an industrial scenario with a collaborative robot where speech phrases and gestures are combined to assign tasks to the robot.

\section{Methods and Tools}

\subsection{Perception Tools}
The perception tools utilized in this work are integrated in a common framework for isolated and human-robot collaborative tasks.
For human perception, Lightweight OpenPose, a human skeleton detection tool \cite{Osokin2018} is used, which takes images (RGB) as input and returns skeleton node points as output. For interaction, the wrist node of the skeleton is taken and, when presented in a certain image area, serves as trigger for robot actions (e.g., stop, continue) or refers to certain objects in the scene (i.e., detected objects pointed to). In the latter case, the detected object that is closest to the wrist node is selected for robot action execution.
Speech recognition is enabled by Vosk \cite{alphacepheiVosk} for the detection of pre-defined input commands and phrases. 
This set of words and sentences relate to available actions of the robot and locations in the scene, as described in Table \ref{tab:commands}. 
The model is configured by filtering out unnecessary words that are unsuitable for robot instructions.
Objects in the scene are detected by a neural network (Detectron2 \cite{wu2019detectron2}) trained on a custom dataset collected for the use case \cite{Sharma2023}. 

\subsection{Multi-modal Perception Methods}
The perception tools can be used in different ways to allow for sensor redundancy, sensor multi-modality and sensor information fusion, as follows.
\begin{itemize}
    \item \textbf{Sensor redundancy} - multiple sensors are used to command the same robot actions, e.g., speech or hand gesture to stop robot motion
    \item \textbf{Sensor multi-modality} - different sensor modalities are used to command individual robot actions, e.g., speech provides the robot actions, vision detects human gestures
    \item \textbf{Sensor-fusion} - different sensor modalities are combined to command a single robot action, e.g., speech provides robot action, vision provides specific object location as pointed to by the human
\end{itemize}

While sensor redundancy and multi-modality is supported and demonstrated in Section \ref{sec:results}, we emphasize our contributions to the fusion of multiple sensor outputs into a single robot command, as explained in the following section.

\begin{table*}[t]
\centering
\caption{Perception methods' input and output}
\label{tab:commands}
\begin{tabular}{p{0.1\textwidth}|p{0.6\textwidth}|p{0.22\textwidth}}

\textbf{Method} &\textbf{Input}       & \textbf{Output}              \\ \hline
\begin{tabular}[c]{@{}l@{}}\textbf{Wrist}\\\textbf{detection}\end{tabular}  & \begin{tabular}[c]{@{}l@{}}RGB image of the scene (human front-facing) \\Human gesture by moving wrist to certain image location\end{tabular} & \begin{tabular}[c]{@{}l@{}}Robot stop/continue actions\end{tabular} \\ \hline
\begin{tabular}[c]{@{}l@{}}\textbf{Speech}\\\textbf{recognition}\end{tabular} & \vspace{-7mm}Robot action commands: \texttt{<pick, place, give, go, stop, pause, continue>} \newline Workspace commands: \texttt{<rod, home, arm, me>}\newline  Human speech requests: \texttt{<place rod>}, \texttt{<go home>}, \texttt{<give me another rocker arm>}, \texttt{<pick up the last rod>} & \begin{tabular}[c]{@{}l@{}}Robot motion\\Gripper actions\\Robot to human hand-over\\Robot stop/continue actions \end{tabular}\\  \hline
\begin{tabular}[c]{@{}l@{}}\textbf{Object}\\\textbf{detection}\end{tabular}  & RGB image of the scene (top-down)  & \begin{tabular}[c]{@{}l@{}}Detected objects in the scene \\Valid target location for robot\end{tabular}\\ \hline
\begin{tabular}[c]{@{}l@{}}\textbf{Co-speech}\\\textbf{gesture}\end{tabular}  & \vspace{-5mm} \texttt{<pick rod>} + pointing gesture + object detection \newline \texttt{<give me this rod>} + pointing gesture + object detection\newline \texttt{<give me that rocker arm>} + pointing gesture + object detection&\begin{tabular}[c]{@{}l@{}}Robot motion\\Gripper actions\\Robot to human hand-over\end{tabular}\\ \hline
\end{tabular}
\end{table*}

\subsection{Co-speech Gesture Model}
The single-modal visual and speech perception models are fused into a multi-modal perception model by combining speech commands, pointing gestures and object detection (see Fig. \ref{fig:co-speech gesture model}). Several examples of these co-speech gestures are described in Table \ref{tab:commands}. The human can refer to individual objects in the scene by speech (e.g., \texttt{<rod>, <rocker arm>}) and pointing to them, and apply specific robot actions by speech commands (e.g., picking with \texttt{<pick>}, placing with \texttt{<place>}, robot to human hand-over with \texttt{<give>}). 

Depending on the object, different robot actions are possible, as specified beforehand. For example, objects can be picked up from the table, placed in specified locations and handed over to the person. Object detection returns a list of objects in the scene, which can be verbally referred to by their class. Pointing gesture detection allows to refer to specific objects in the scene by relating the pointing gesture location to detected object locations. Robot actions are therefore commanded by specific action verbs and object classes, complimented by gestures to provide fine-grained object references (see Fig. \ref{fig:co-speech gesture model}).

\section{Experimental Results}\label{sec:results}
\subsection{Industrial Use Case}
The considered use case replicates an industrial assembly task that in current situation is done manually by human operators. The solution we propose introduces a collaborative robot as assistive tool to the assembly station, under control of the person. This means that the assembly work is coordinated by the human, with the robot assisting in tasks that the human decides. Available robot actions are to move to certain locations in the work space, pick objects that are detected on the table, place objects to specified locations or hand them over to the human. In addition, coordinated actions include the stopping and continuing of robot actions during execution, for human visual inspection of the objects placed by the robot. Human commands can be communicated by hand gestures and/or speech, with different levels of functionality as described in Table \ref{tab:commands}. The setup for experiments is depicted in Fig. \ref{fig:setup} and includes two cameras (Intel Realsense D435) for visual perception (one front-facing for wrist detection; Fig. \ref{fig:setup}(b) and one top-down for object detection; Fig. \ref{fig:setup}(c)) and a microphone for speech recognition. Computation is performed on a standard Desktop PC running Ubuntu Linux with Nvidia GTX 1080 Ti GPU, and all robot (Franka Emika) communication and control utilizes ROS. All tools are open-source available to utilize or replicate: \url{https://github.com/opendr-eu/opendr}.


\subsection{Human Gesture Detection}
Results for the visual wrist detection tool are depicted in Fig. \ref{fig:setup}(b), which highlights both detected human wrists. When one of the wrists is detected inside one of the squares, this is taken as trigger for referring to certain robot actions or objects in the scene. For example, to stop robot motion, the left wrist should be detected in the top left square and to continue robot motion, the right wrist should be detected in the top right square. Pointing gestures are interpreted in a similar manner. When the human points to a certain object, first the left or right wrist needs to be detected in either of the lower two squares in the image, after which the location to the closest detected object is determined. Performance of the skeleton detection tool has been reported in the original paper \cite{Osokin2018}. In our use case the detection accuracy of the wrists inside a square is consistent around 90\%, as assessed from 20-second interval tests for different squares. This is satisfactory for effective collaboration. 

\subsection{Object Detection}
\begin{figure*}[t!]
\centering
\subcaptionbox{}{%
  \includegraphics[height=0.30\linewidth]{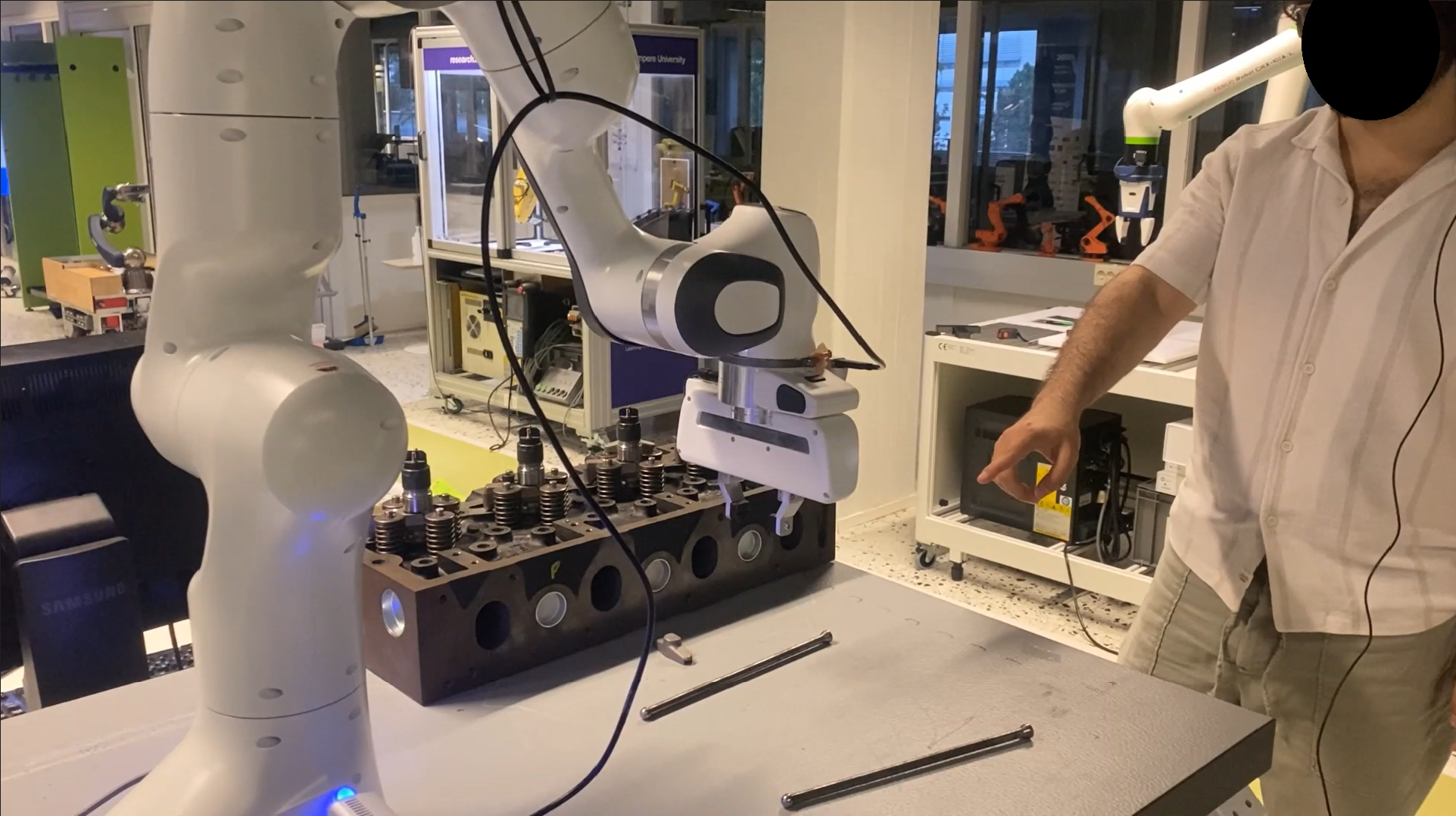} 
}\hfill 
\subcaptionbox{}{%
  \includegraphics[height=0.30\linewidth]{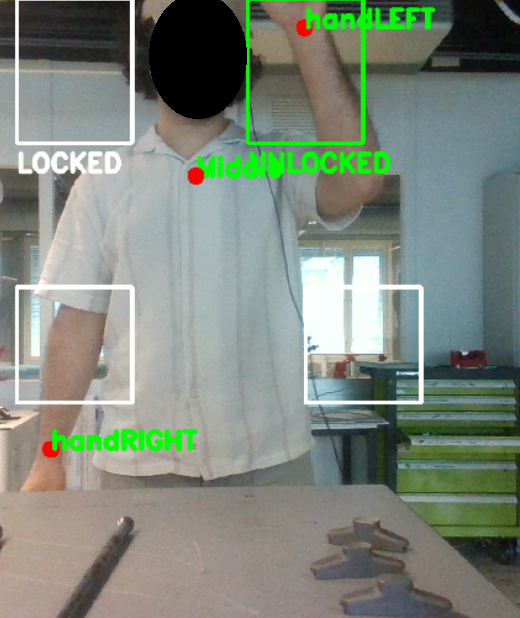} 
}\hfill 
\subcaptionbox{}{%
  \includegraphics[height=0.30\linewidth]{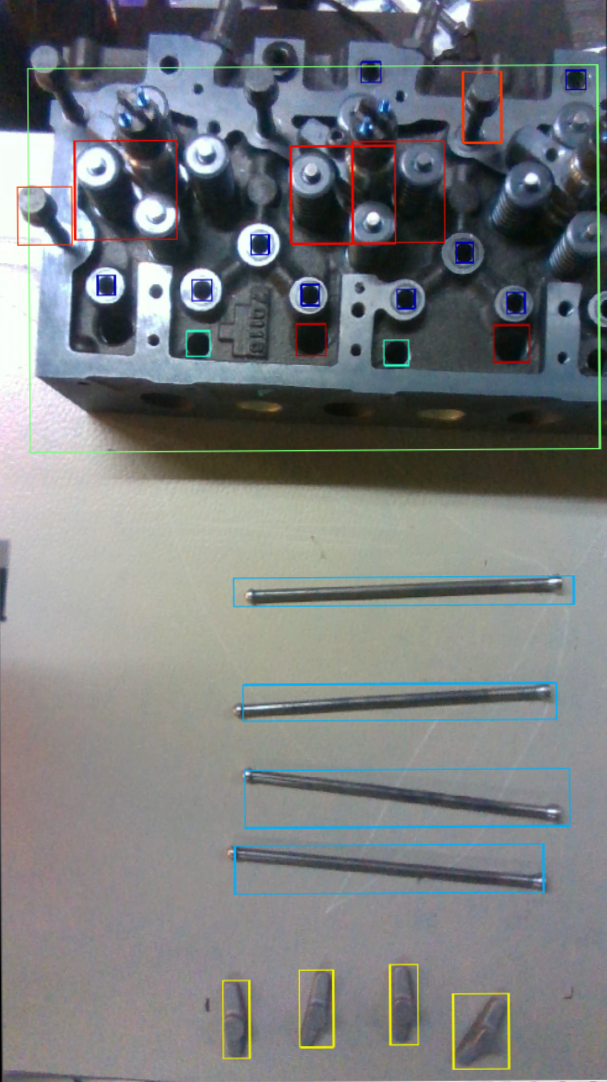} 
}\hfill 
  \caption{Experimental setup with a human pointing at an object for robot picking (a). One camera is human front-facing to capture human hand gestures (b), one camera is mounted on the robot (eye-in-hand) for object detection on the table (c).\label{fig:setup}}
\end{figure*}

Results of visual object detection are depicted in Fig. \ref{fig:setup}(c), which has the different detected objects annotated by colored bounding boxes (yellow for the rocker arms and blue for the rods). As objects are detected in image space, careful calibration of both cameras ensures the detected objects can be picked from the table and that pointing gestures can refer to the same object in both camera frames. In our use case the detection accuracy of all classes is over 90\%. 

\subsection{Speech Recognition}
Results of speech recognition were found satisfactory, as in most cases the spoken commands are recognized correctly. Performance, as reported in \cite{alphacepheiVosk}, depends on the language skills of the person giving commands, as in certain cases non-native English speakers had to speak more clear to achieve correct speech recognition. Besides the speech recognition itself, the speech tool was improved by including a voice activity detector and a time-delay filter (0.5 seconds) to consider the natural pause in human speech. This resulted in a delay of $\approx1.9$ seconds between a verbal command and the recognized speech (average of 50 trials with different commands).

\subsection{Co-speech Gesture Model Performance}
The co-speech gesture model has all three perception models running in parallel, decreasing slightly the running performance of the skeleton detection tool (i.e., 24 fps with image size of 1920$\times1080$). Object detection achieves a frame rate of 4.5 fps with image size of 1280$\times$720.
Extended experiments were performed to test the co-speech tool in a collaborative assembly scenario. This included a human and robot performing assembly steps to an engine, with parts that are either mounted by the person or by the robot.
Parts assembled by the person are picked by the robot from the table and handed over to the human, and parts assembled by the robot are picked by the robot from the table and directly mounted to the engine. Coordination of the tasks and requesting robot actions is done by the person via the co-speech gesture model. In addition, the human can halt and continue robot tasks at any time, by both gesture and speech commands (\texttt{<stop>}, \texttt{<pause>}, \texttt{<continue>}).

\begin{figure*}[!t]
  \centering
\subcaptionbox{}{%
  \includegraphics[width=0.32\textwidth]{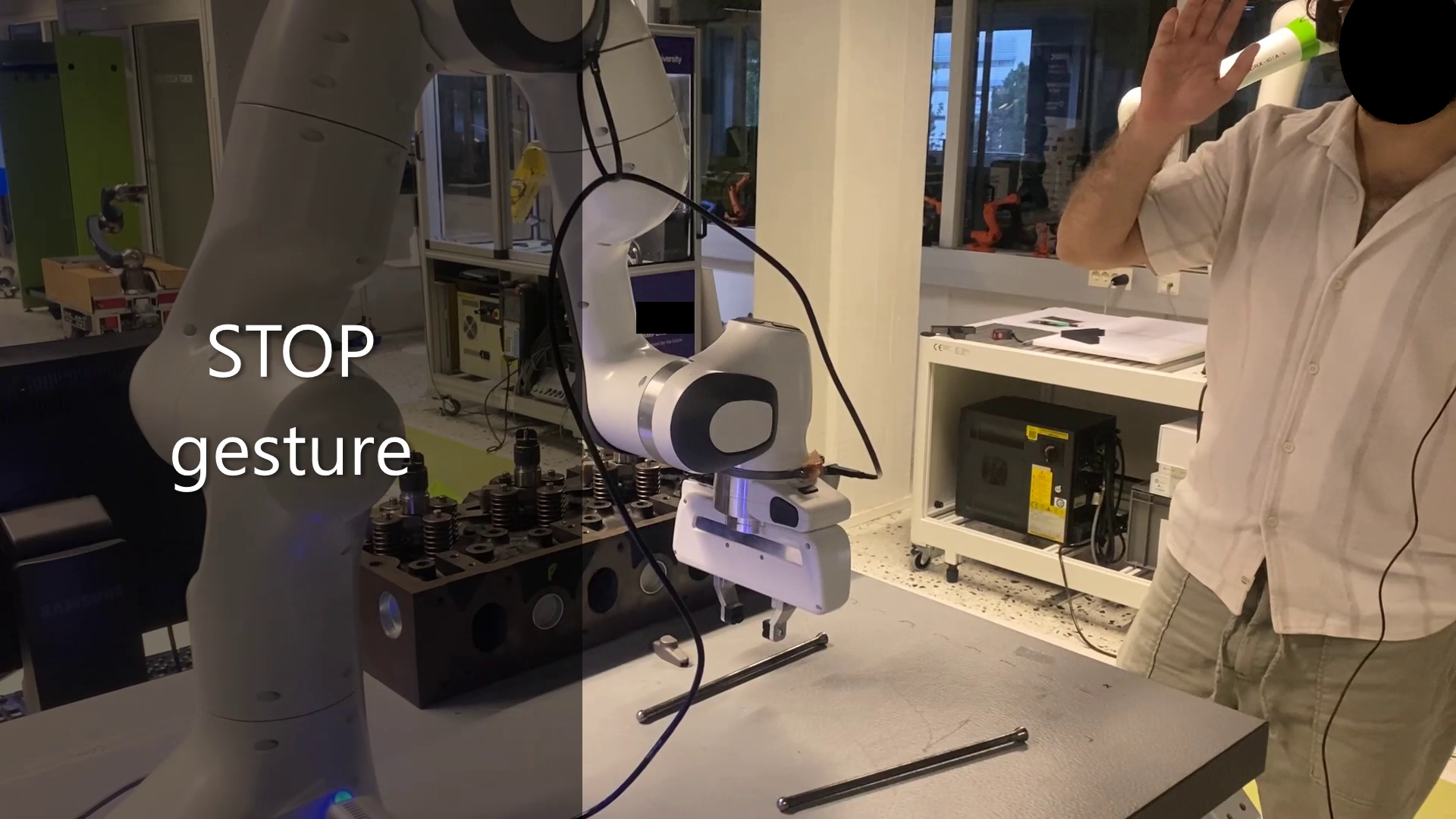} 
}\hfill 
\subcaptionbox{}{%
  \includegraphics[width=0.32\textwidth]{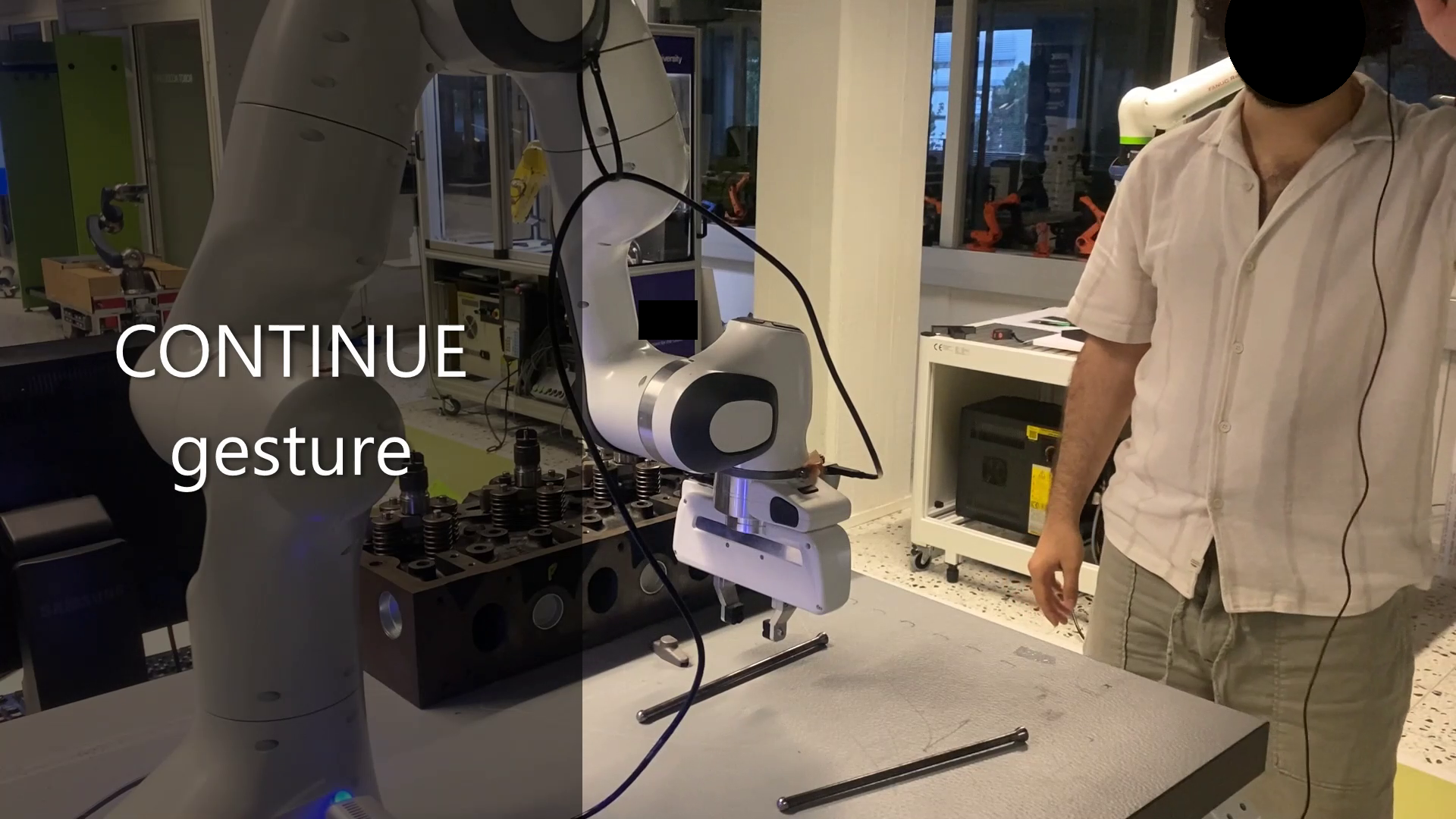} 
}\hfill 
\subcaptionbox{}{%
  \includegraphics[width=0.32\textwidth]{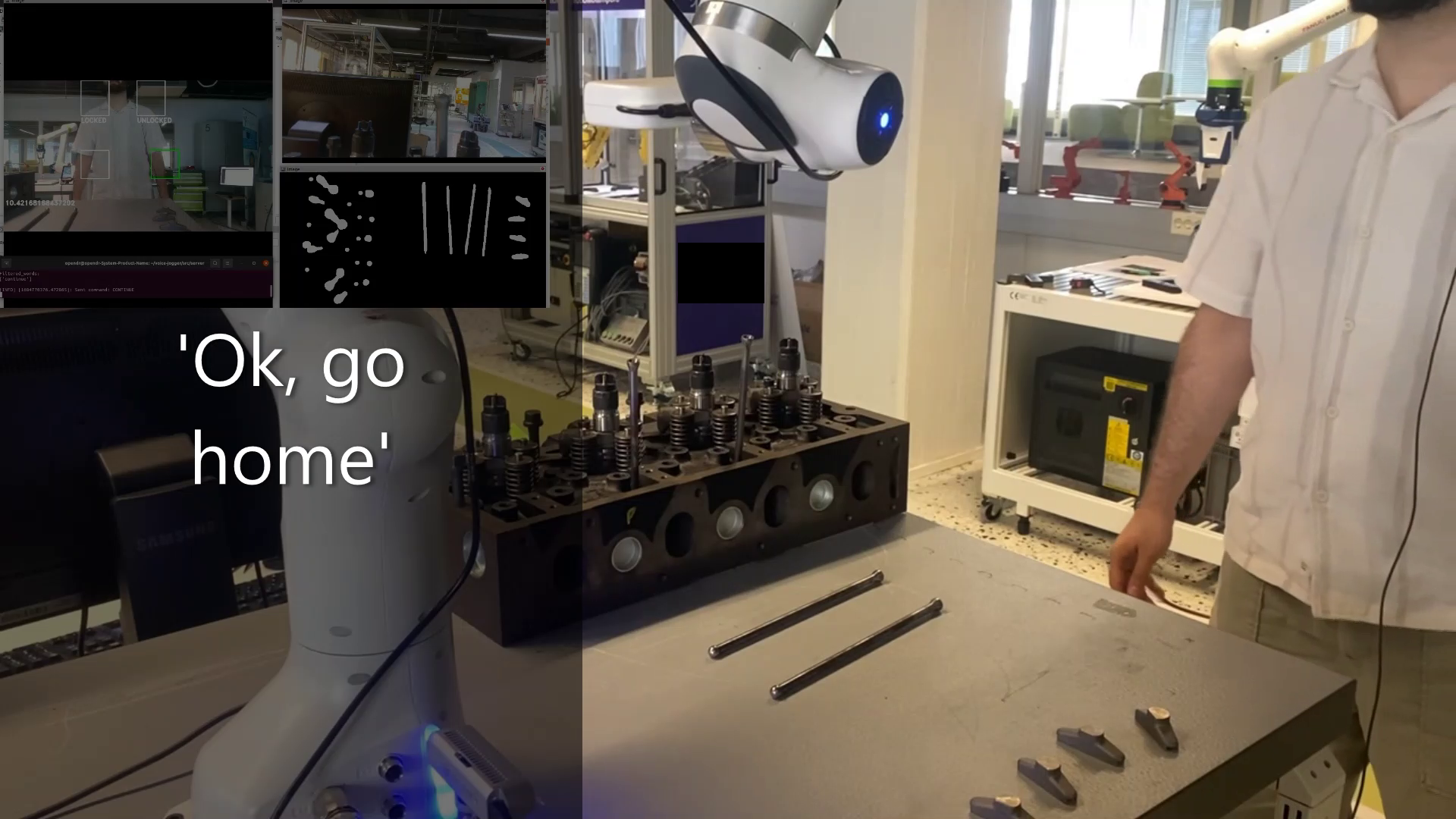} 
}\hfill 
\caption{Single command gestures stop (a), continue (b) and speech (c). \label{fig:single_commands}}

  \centering
\subcaptionbox{}{%
  \includegraphics[width=0.32\textwidth]{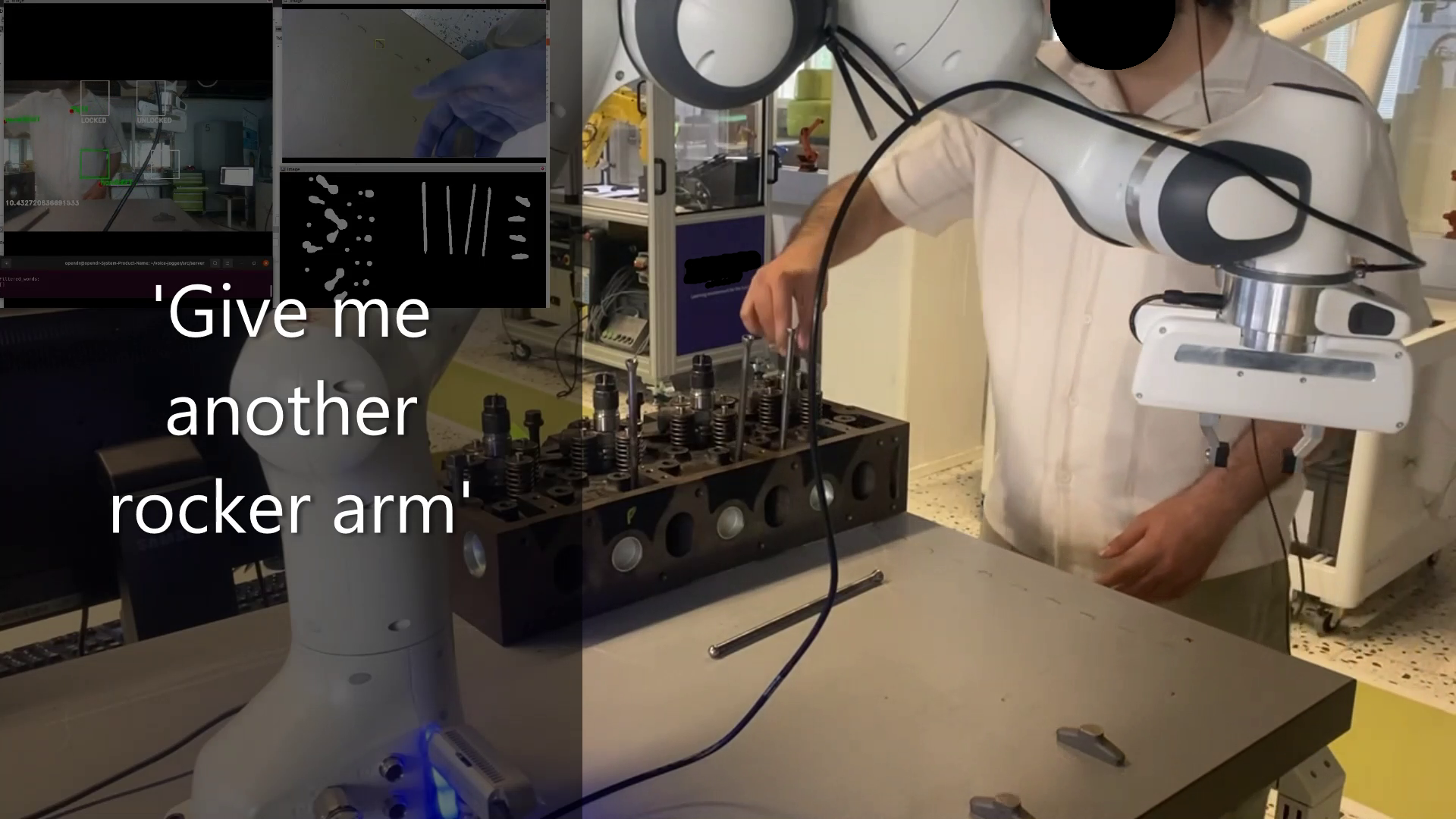} 
}\hfill 
\subcaptionbox{}{%
  \includegraphics[width=0.32\textwidth]{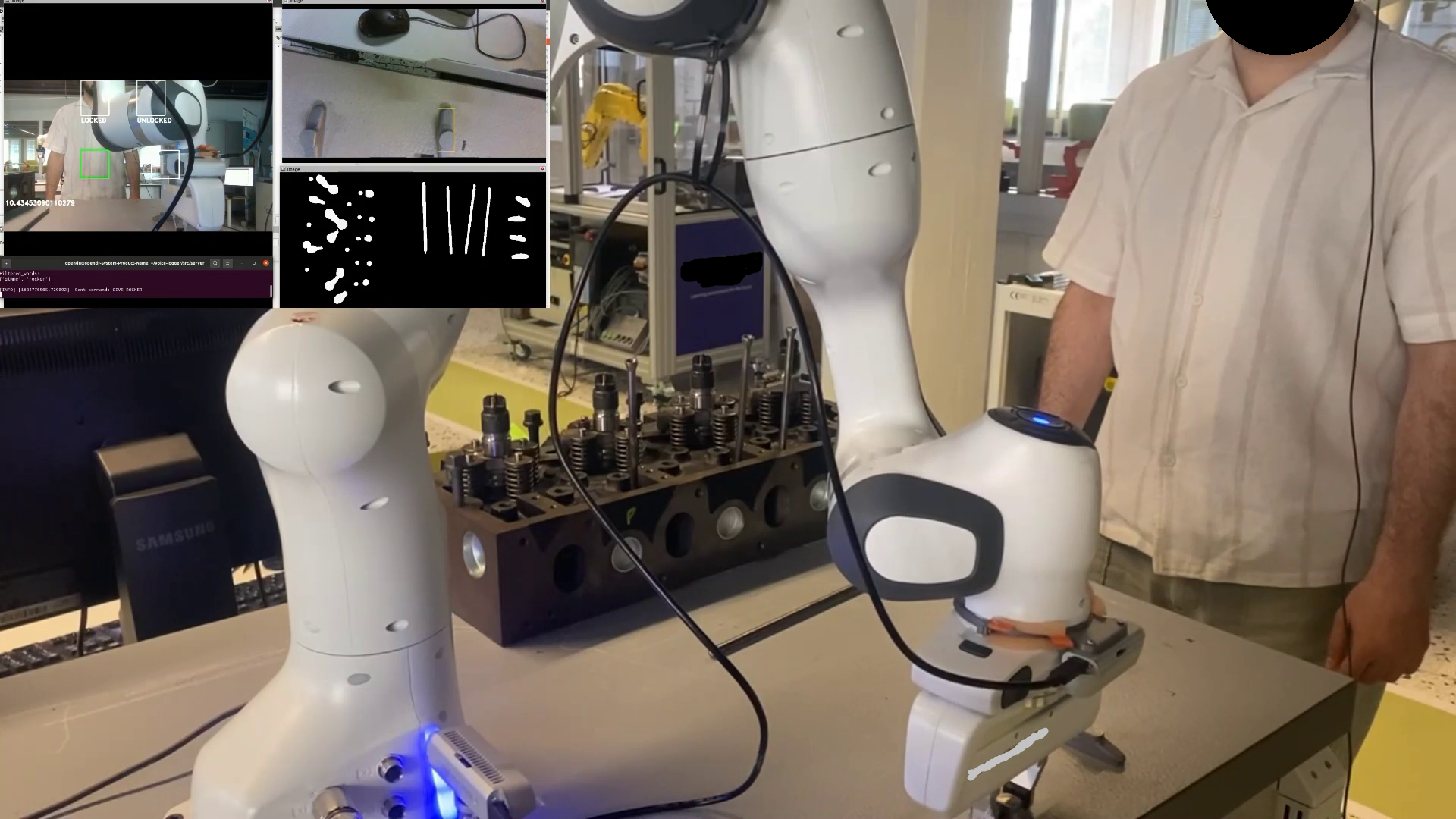} 
}\hfill 
\subcaptionbox{}{%
  \includegraphics[width=0.32\textwidth]{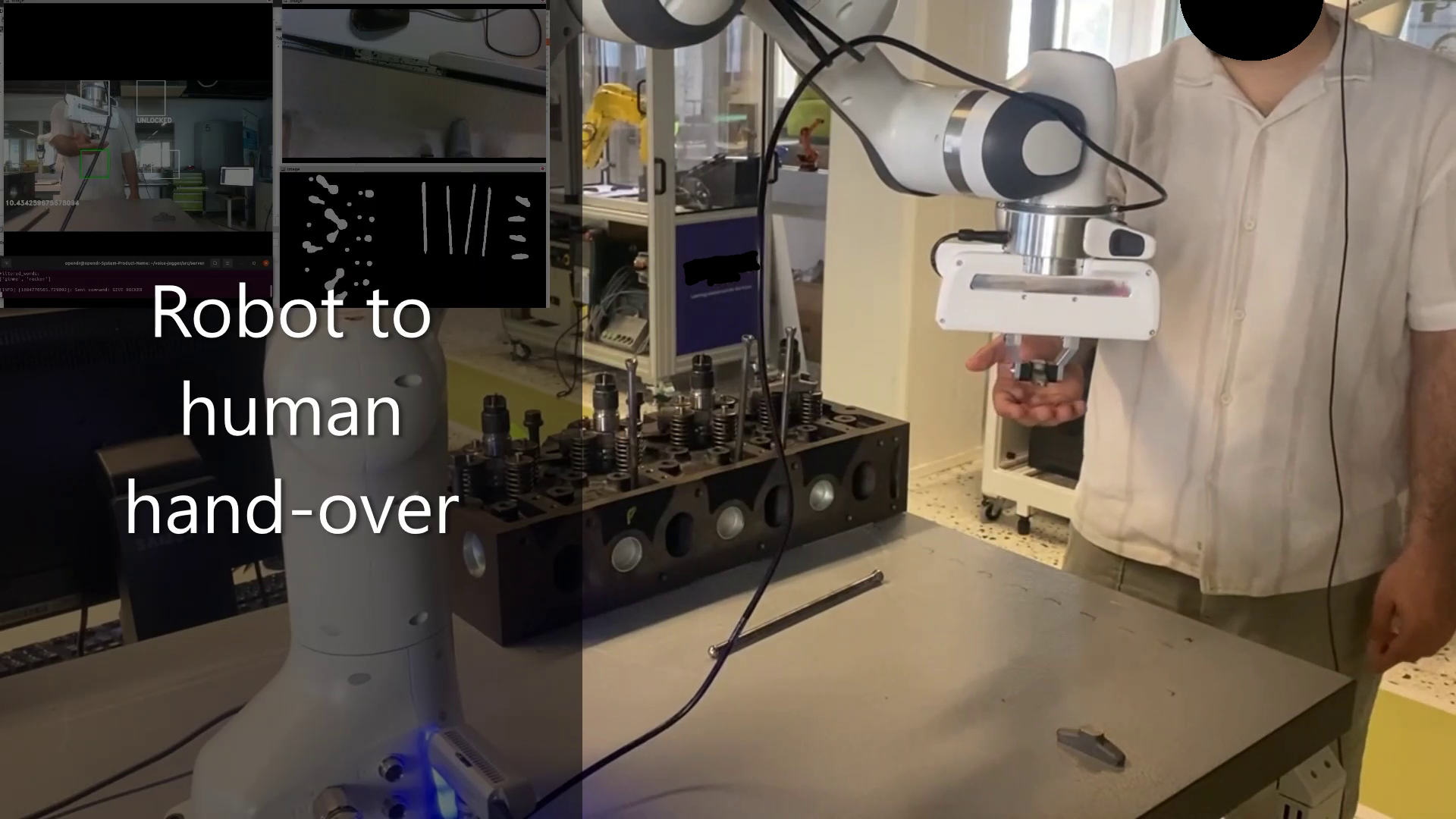} 
}\hfill 
\caption{Speech phrase to achieve robot to human hand-over. \label{fig:command_refer}} 

  \centering
\subcaptionbox{}{%
  \includegraphics[width=0.32\textwidth]{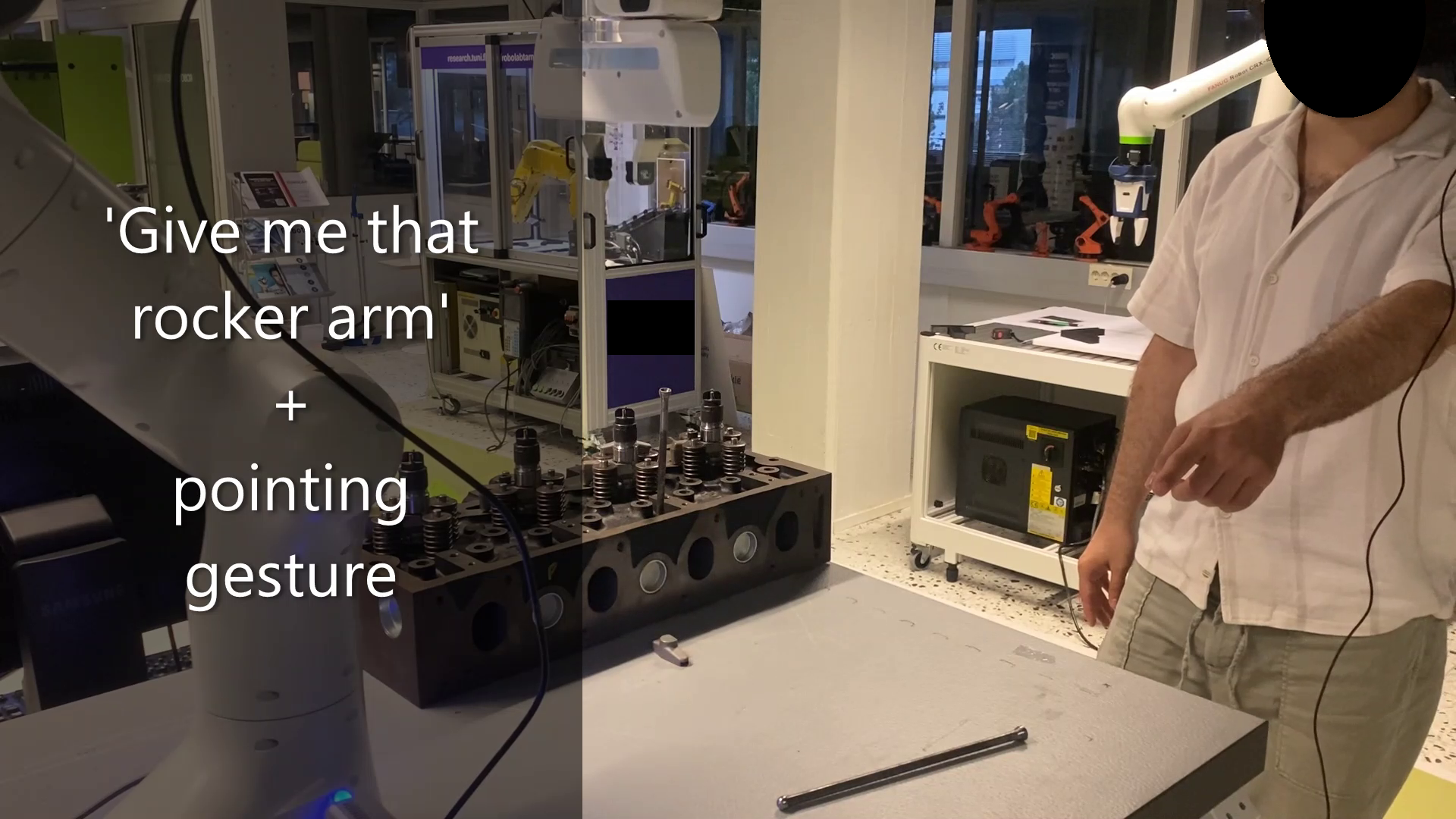} 
}\hfill 
\subcaptionbox{}{%
  \includegraphics[width=0.32\textwidth]{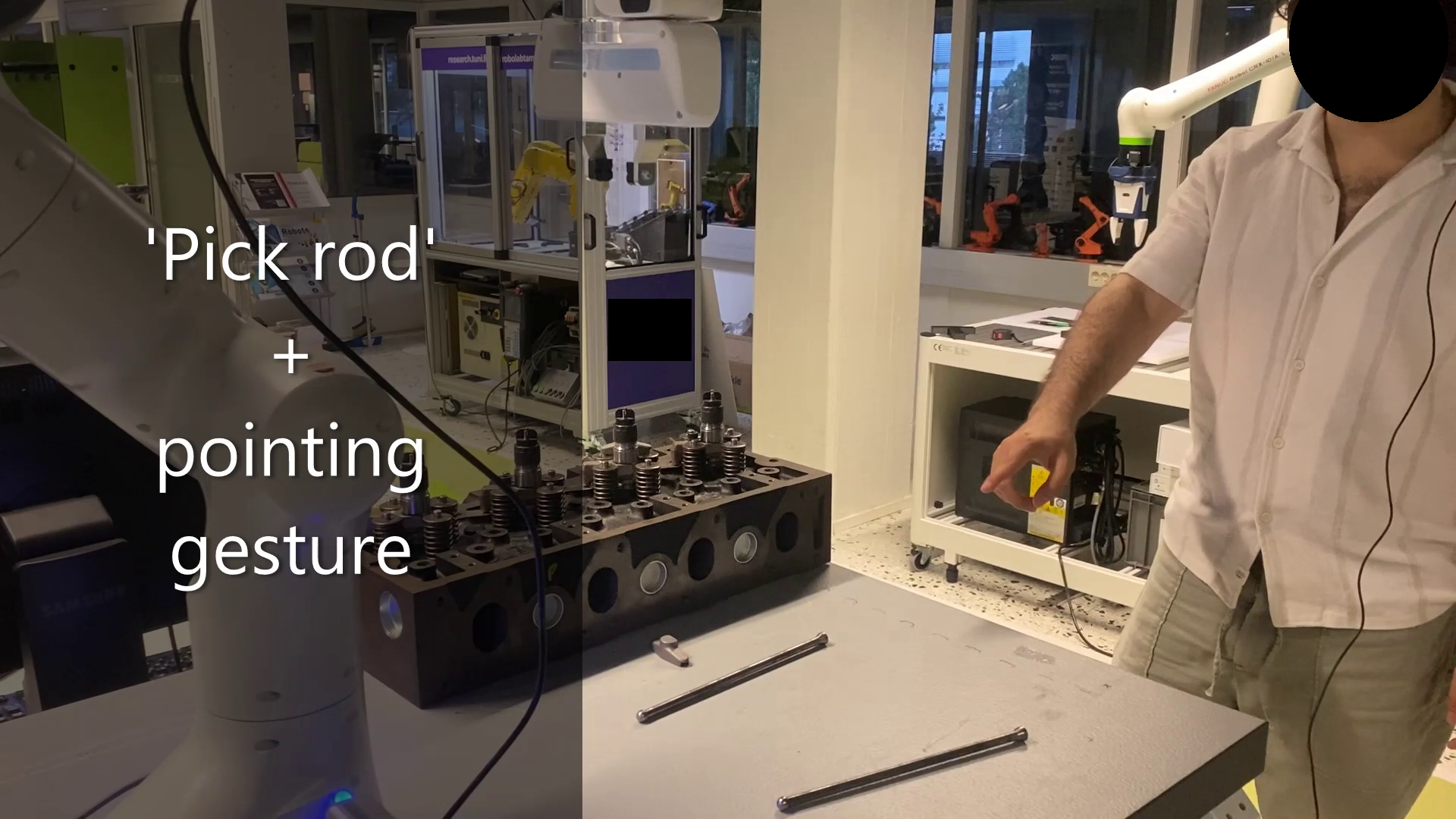} 
}\hfill 
\subcaptionbox{}{%
  \includegraphics[width=0.32\textwidth]{figs/GIVE_rod_pointing_anon.png} 
}\hfill 
\caption{Co-speech gestures to achieve specified robot actions to objects. \label{fig:co-speech_command}}
\end{figure*}

\textbf{\textit{Single commands -}}
Fig. \ref{fig:single_commands}(a-b) depict the human commanding a stop and continue gesture, respectively. Fig. \ref{fig:single_commands}(c) shows the human commanding the robot to move to its 'home' configuration by the phrase \texttt{<ok, go home>}. For this, the home location is preprogrammed in the software scripts.

\textbf{\textit{Speech phrases -}}
Fig. \ref{fig:command_refer} depicts how human speech alone can be utilized to command robot actions, by the phrase \texttt{<give me another rocker arm>}. From the recognized speech, the tool extracts relevant words and connects these to robot actions and objects in the scene. In this case \texttt{<give me>} refers to a robot to human hand-over, \texttt{<rocker arm>} refers to the rocker arm class in the object detection model, and \texttt{<another>} implies any of the detected rocker arms, meaning the first in the returned detection list. As a result, the command phrase initiates all required robot actions and starts executing them one-by-one, as shown in Fig. \ref{fig:command_refer}(ac).

\textbf{\textit{Co-speech commands -}}
Fig. \ref{fig:co-speech_command} depicts examples of the co-speech gesture model that utilizes a human speech phrase and pointing gesture to achieve robot actions applied to specified objects in the scene. In this case, as a pointing gesture is detected by the wrist detection tool, the closest specified object to the human wrist is selected for the robot actions. A video of the co-speech gesture model demonstrates all commands from Fig. \ref{fig:single_commands}-\ref{fig:co-speech_command}: \url{https://youtu.be/b_ISrhOlcC8}. This shows the collaborative tasks, where the human coordinates the actions of the robot with four pick and place actions and four robot to human hand-overs. Human inspection is done after object placement by stopping robot motion with a speech command. In total, the experiment includes over 20 speech commands and seven co-speech gestures to coordinate the shared task. 

\section{Discussion and Limitations}
Sensor redundancy enables different modalities to command the same robot action. This was demonstrated for stopping and continuing robot motion and actions by hand gestures (see Fig. \ref{fig:single_commands}) and by speech commands. 
While hand gestures can be detected at relatively high rate (>24 FPS), it can takes several image frames before a correct prediction occurs. On the other hand, speech commands can have considerable delay even when a first verbal command is correctly recognized. 



 While in most cases the co-speech gesture model achieves the intended robot commands and collaboration, some limitations are identified. First, detection of the human wrist in a specific image location requires careful human hand motion. As alternative, human hand gestures could be recognized directly from a dedicated model \cite{Mazhar2019}. In our case, inference time and detection accuracy were the main reasons for utilizing a skeleton detection model instead.
 Second, the relation between human pointing and objects in the scene needs precise camera calibration, such that the same object is referred to in both images. This can be circumvented by using a single camera for both visual perception tools, with RGB and depth perception functionalities.

\section{Conclusions}
This work investigated how multiple perception tools can be utilized and combined for effective human-robot collaboration. Human hand gestures and speech, as well as object detection, provide the input for robot actions, as coordinated by a person. Single modal perception serves to command basic robot actions (stop, continue) by gesture or speech. A co-speech gesture model is developed that combines human speech phrases, pointing gestures and object detection to command robot actions (pick and place, robot to human hand-overs) to specified objects in the scene. Experimental results demonstrate that co-speech gestures can be easily utilized for coordinating a shared task between human and robot. 



\bibliographystyle{IEEEtran}
\bibliography{IEEEabrv,refs}

\end{document}